\documentclass{article}

\usepackage{arxiv}
\usepackage[utf8]{inputenc} 
\usepackage[T1]{fontenc}    
\usepackage{hyperref}       

\usepackage{url}            
\usepackage{booktabs}       
\usepackage{amsfonts}       
\usepackage{nicefrac}       
\usepackage{microtype}      
\usepackage{lipsum}		
\usepackage{graphicx}
\usepackage{natbib}
\usepackage{doi}
\usepackage{microtype}
\usepackage{graphicx}
\usepackage{caption}
\usepackage{subcaption}
\usepackage{booktabs} 

\usepackage{amsmath}
\usepackage{amssymb}
\usepackage{mathtools}
\usepackage{amsthm}
\usepackage{cleveref}

\title{PARCv2: Physics-aware Recurrent Convolutional Neural Networks for Spatiotemporal Dynamics Modeling}

\author{ Phong C.H. Nguyen \\
	School of Data Science\\
	University of Virginia\\
    Charlottesville, VA 22903 \\
	\And
	Xinlun Cheng \\
	School of Data Science,\\
	Department of Astronomy\\
	University of Virginia \\
	Charlottesville, VA 22903 \\
    \And
    Shahab Azarfar   \\
    School of Data Science\\
	University of Virginia\\
    Charlottesville, VA 22903 \\
    \And
    Pradeep Seshadri   \\
    Department of Mechanical Engineering\\
    University of Iowa \\
    Iowa City, IA 52242 \\
    \And
    Yen T. Nguyen  \\
    Department of Mechanical Engineering\\
    University of Iowa \\
    Iowa City, IA 52242 \\
    \And
    Munho Kim   \\
    School of Mechanical Engineering \\
    Kyungpook National University	\\
    Daegu, Republic of Korea \\
    \And
    Sanghun Choi  \\
    School of Mechanical Engineering \\
    Kyungpook National University	\\
    Daegu, Republic of Korea \\
    \And
    H.S. Udaykumar  \\
    Department of Mechanical Engineering\\
    University of Iowa \\
    Iowa City, IA 52242 \\
    \And
    Stephen Baek \thanks{Corresponding author: baek@virginia.edu}  \\
    School of Data Science\\
    Department of Mechanical and Aerospace Engineering \\
	University of Virginia\\
    Charlottesville, VA 22903 \\
}

\date{}

\renewcommand{\shorttitle}{\textit{arXiv} Template}

\hypersetup{
pdftitle={A template for the arxiv style},
pdfsubject={q-bio.NC, q-bio.QM},
pdfauthor={David S.~Hippocampus, Elias D.~Striatum},
pdfkeywords={First keyword, Second keyword, More},
}

\begin{document}
\maketitle

\begin{abstract}
	Modeling unsteady, fast transient, and advection-dominated physics problems is a pressing challenge for physics-aware deep learning (PADL). The physics of complex systems is governed by large systems of partial differential equations (PDEs) and ancillary constitutive models with nonlinear structures, as well as evolving state fields exhibiting sharp gradients and rapidly deforming material interfaces. Here, we investigate an inductive bias approach that is versatile and generalizable to model generic nonlinear field evolution problems. Our study focuses on the recent physics-aware recurrent convolutions (PARC), which incorporates a differentiator-integrator architecture that inductively models the spatiotemporal dynamics of generic physical systems. We extend the capabilities of PARC to simulate unsteady, transient, and advection-dominant systems. The extended model, referred to as PARCv2, is equipped with differential operators to model advection-reaction-diffusion equations, as well as a hybrid integral solver for stable, long-time predictions. PARCv2 is tested on both standard benchmark problems in fluid dynamics, namely Burgers and Navier-Stokes equations, and then applied to more complex shock-induced reaction problems in energetic materials. We evaluate the behavior of PARCv2 in comparison to other physics-informed and learning bias models and demonstrate its potential to model unsteady and advection-dominant dynamics regimes.
\end{abstract}


\section{Introduction}
Physics-aware deep learning (PADL) has been gaining attention for the simulation of spatiotemporal dynamics, \textit{i.e.}, time-evolving fields of physical systems. The physical behaviors of continua are represented as a system of partial differential equations (PDE), expressing the rate of change $\frac{\partial \mathbf{x}}{\partial t}$ of a field $\mathbf{x}(t, \mathbf{r})$ as a function $f(t, \mathbf{x}, \mathbf{y}, \nabla\mathbf{x}, \nabla\mathbf{y}, \cdots)$ where $t$ is time, $\mathbf{r}\in \mathbb{R}^d$ is a position vector, $\mathbf{x}$ is the evolving field variable, and $\mathbf{y}(t,\mathbf{r})$ is a covariate field that influences the evolution of $\mathbf{x}$. As a simple case, the heat equation $\frac{\partial \mathbf{x}}{\partial t} = \nabla^2 \mathbf{x}$ is one such governing PDE that describes the diffusion of heat and appears frequently in various computer vision and machine learning literature.

As the complexity of the physics displayed by a system increases, the governing PDEs increase in number and complexity accordingly, with nonlinear coupling between the dependent variables and source terms inducing a wide range of spatial and temporal scales. The constitutive laws describing the material response can also become complex nonlinear differential equations. Furthermore, depending on the physics problem, governing PDEs relating $\mathbf{x}$ and $\mathbf{y}$ may not even be available in an explicit form, and hence the equations governing the evolution of these variables may need to be learned from data. 

Particularly, interesting machine-learning challenges may emerge when there are large, localized spatial gradients in $\mathbf{x}$ and $\mathbf{y}$. Examples may include shocks, fast transients, collapsing interfaces, reaction fronts, and evolving internal/external boundaries interacting with the evolving field. An archetype of such problems is the shock-induced initiation of porous energetic materials \citep{Perry2018,Mang2021,Mi2020}, as in the benchmark problem presented in Section \ref{sec:exp_em}. When exposed to external shock loading, pores in these materials collapse and release strong localized energy, forming high-temperature regions called ``hotspots.'' In this process, both the traveling shock and the energy localization near hotspots render intense spatiotemporal gradients in the evolving temperature and pressure fields. Collapsing pore boundaries also interact with these evolving fields, producing intricate sharp patterns on the reaction front. Hence, accurately predicting such sharp and complex features evolving over time is critical to capture the physics of ignition and reaction in shocked reactive materials.

Simulating the intricate physics of systems with fast transients and evolving spatiotemporal fields demands high-fidelity numerical techniques, incurring substantial computational costs and resulting in limited training data for machine learning. To address this challenge, previous works have used the squared sum of residuals of PDEs, known as \textit{physics-informed loss}, to regularize the training. However, strong regularization can introduce undesired ``smoothing effects,'' generate conflicting network gradients during training, and may be inapplicable when direct mathematical relationships between the evolving field $\mathbf{x}$ and covariate fields $\mathbf{y}$ are unavailable. Given these practical scenarios, our goal is to model the governing PDEs of spatiotemporal dynamic systems using neural networks. This enables us to either approximate the solution of a complex PDE system (Sections \ref{sec:exp_burgers} $\&$ \ref{sec:exp_ns}) or discover unknown governing PDEs from data (Section \ref{sec:exp_em}).

We hypothesize that a previously published inductive modeling approach called \textit{physics-aware recurrent convolution}, or \textit{PARC}, may offer a route to addressing the above challenges \citep{nguyen2023sciadv}. 
PARC presents advantages originating from its recurrent \textit{differentiator-integrator} architecture: governing PDEs of generic dynamics systems are learned from data by the \textit{differentiator network} and then integrated to predict the future states of the evolving field by the \textit{integrator network}.
The previous work of \citeauthor{nguyen2023sciadv} demonstrated the predictive capabilities of PARC when applied to the problem of predicting hotspot evolution in shocked porous reactive materials. Additionally, the physics-awareness aspect of PARC was also justified to a certain extent. 
However, the validation study in the previous work was limited to a single problem obtained by regularizing the advection phenomena and the model's predictive capability was not thoroughly scrutinized on more generic problems.

In this study, we first substantively extend the work of \citet{nguyen2023sciadv} by introducing new design considerations and training schemes for PARC, enabling it for unsteady, fast transient, and advection-dominant field evolution problems. The extended PARC, referred to as \textit{PARCv2}, incorporates spatial derivatives to model more general advection-diffusion-reaction equations. Additionally, PARCv2 utilizes a hybrid integration method, leveraging both numerical and data-driven integration for more stable and accurate predictions for unsteady and highly transient physics problems. 

Furthermore, we examine and characterize the behavior of PARCv2 on standard fluid dynamics benchmark problems, namely \textit{Burgers} and \textit{Navier-Stokes} equations, testing its prediction accuracy and compliance with the known governing PDEs. These benchmark problems represent simple diffusion-dominant systems with the weak influence of advection (Burgers) and advection-dominant fluids with boundaries and unsteady vorticity patterns (Navier-Stokes). Finally, we test PARCv2 on a real-world physics problem of predicting the evolving temperature and pressure fields in shocked energetic materials. This problem exhibits strong advection dominance, fast transient reactions, and sharp spatiotemporal gradients, challenging PARCv2's limits on its predictive power.


\begin{figure*}[tbh!]
     \centering
     \includegraphics[width=0.65\textwidth]{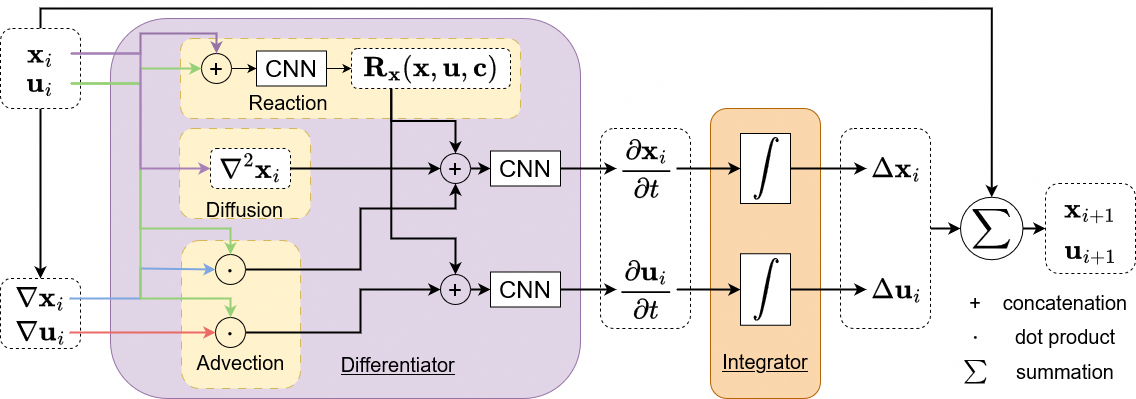}
\caption{PARCv2 architecture.
}
\label{fig:architecture}
\end{figure*}

\section{Related Works}
\label{sec:related_works}
According to the taxonomy proposed by \citet{Karniadakis2021}, there are three different ways to inform machine learning models with prior physics knowledge. One of these three categories is the \textit{observational bias}, which incorporates physics principles and constraints implicitly through data \citep{Cheng2023, Yang2019a}. Observational bias approaches are straightforward to implement with off-the-shelf deep learning models. However, to generate predictions that adhere to the necessary physical laws they require sufficiently large sets of data, which are usually unavailable in the physical sciences.

\paragraph{Learning Bias Methods} On the other hand, \textit{learning bias} approaches, incorporate the governing PDEs more explicitly into the training process. A representative body of work in this category is exemplified by the \textit{physics-informed neural networks}, or \textit{PINN} \citep{Raissi2019}. PINN directly approximates the solution space of PDEs using neural networks. Derivative terms in PDEs are computed using automatic differentiation \citep{JMLR_AD}. The \textit{physics-informed loss} term guides the network parameters during training to minimize the PDE residuals. \citet{Raissi2019} demonstrated the efficacy of such formulation on Navier-Stokes equation, Allen–Cahn equation, Schr\"odinger equation, and other benchmark problems in physics.


The initial success of \citet{Raissi2019} motivated vigorous research centered around PADL  in recent years. Numerous PADL models branched out of PINN, such as PhyCRNet \citep{ren2022phycrnet} and physics-informed neural operators \citep{li2023physicsinformed}. These models were tested and validated on various physics problems, contributing to the accumulation of a significant body of scientific knowledge around PADL. 
From these previous endeavors, the research community has discovered several important machine-learning challenges, especially for physics problems that are unsteady, fast transient, and advection-dominant. 

For instance, problems such as wave equations with high wave number \citep{Basir} or advection flow with a high advection coefficient \citep{krishnapriyan2021characterizing} are known to yield intricate loss landscapes for the physics-informed loss terms (PDE residuals), which hamper the convergence of a model during training \citep{krishnapriyan2021characterizing}. A few solutions were proposed to address this issue, such as casting the physics-informed loss terms as equality constraints \citep{Basir} or adopting the curriculum training strategy \citep{krishnapriyan2021characterizing}. These solutions were proven to smoothen the PINN loss landscape and facilitate its convergence.

Spectral bias is another current challenge, arising from the tendency of PINN models to favor low-frequency modes in field evolution problems. This is largely due to unbalanced sampling of high-frequency patterns during the evaluation of PDE residuals, and can thus be somewhat mitigated by incorporating adaptive weighting \citep{WANG2022110768} or Fourier/spectral features \citep{WANG2021113938}. However, for physics problems with large Kolmogorov $n$-width, \textit{e.g.}, ones exhibiting high advection dominance, spatial localization phenomena, discontinuous phases, and/or sharp gradients, learning bias models still fail to learn crucial intricate details \citep{MOJGANI2023115810}.

\paragraph{Inductive Bias Methods} In contrast to enforcing physical constraints by minimizing PDE residuals, \textit{inductive bias} approaches attempt to embed prior physics knowledge within the network's computational graph itself. For example, equivariant neural networks \citep{Weiler2018, Batzner_2022, Gong_2023, ma2023learning} preserve the symmetry under a group operation (\textit{e.g.}, the standard Euclidean group $SE(n)$) by redefining the convolution filters in such a way that the convolution responses are equivariant under certain geometric transformations.

However, little to no previous work has been done for inductive bias modeling of more general nonlinear field evolution problems that extend beyond problems exhibiting group symmetry. \textit{Finite volume neural network} (FINN), introduced by \citet{karlbauer22a}, is among these minorities. In FINN, multiple neural network modules are used to model specific parts of the target PDE. These modules interact with each other in a distributed and compositional manner, aiming to model the physics governed by the target PDE. FINN demonstrated success in dealing with various types of PDEs, including Burgers, diffusion-reaction, and Allen{–}Cahn equations. However, one of the limitations of FINN is that the PDE constants, \textit{e.g.}, the diffusion coefficient, are treated directly as network parameters (either trainable or untrainable). This restricts the model's expressiveness in modeling nonlinearities in PDEs and may limit its performance when dealing with PDEs with large variations of constants.

The \textit{physics-aware recurrent convolutional neural networks}, or \textit{PARC} \citep{nguyen2023sciadv,nguyen2023pep} is the another notable work in this category. PARC aims to model the governing equation of a simplified dynamical system $\frac{\partial \mathbf{x}}{\partial t} = f(t,\mathbf{x},\mathbf{y},\nabla\mathbf{x},\nabla\mathbf{y},\cdots)$ and its time integration using recursively connected convolutional neural network (CNN) blocks, called \textit{differentiator} and \textit{integrator}. The differentiator CNN approximates the function $f$ in the above expression. The time derivative $\frac{\partial \mathbf{x}}{\partial t}$ predicted by the differentiator CNN is then integrated by the integrator CNN, which approximates the integral operator using a CNN. Applied recursively, the differentiator-integrator architecture emulates the process by which traditional physics solvers obtain the solution of governing PDEs. 


However, despite certain success, the previous work of \citet{nguyen2023sciadv} does not fully extend to a broader spectrum of PDEs, especially the ones in the regime of strong advection dominance. \citeauthor{nguyen2023sciadv} circumvented such a limitation by normalizing the velocity field and removing advection, but it is questionable if such a normalization strategy would generalize to more complex, transient physics phenomena.

\section{Method}
\label{sec:methods}
Here, we consider a wide group of \textit{advection-diffusion-reaction} processes, in which the temporal evolution of a state field $\mathbf{x}$ in a continuum domain is the combination of \textit{advection} $(\mathbf{u} \cdot \nabla) \mathbf{x}$ driven by a physical force represented with a (velocity) vector field $\mathbf{u}$, \textit{diffusion} $(\nabla \cdot k \nabla) \mathbf{x}$ with diffusion coefficient $k$, and the generation/consumption of a source/sink (\textit{reaction}) $\mathbf{R}(\mathbf{x},\mathbf{c})$. By explicitly incorporating the spatial gradient $\nabla \mathbf{x}$ of the state field and its divergence $\nabla \cdot \nabla \mathbf{x}$ into the differentiator architecture, we inform PARCv2 with physics priors (Section \ref{sec:arch}). Furthermore, a hybrid integration approach is introduced to leverage the benefits of numerical integration with data-driven integration (Section \ref{sec:data_driven_int}), enabling accurate and stable predictions over a long time period.
Finally, we also present a two-stage training scheme that separates the process of learning the PDEs and the time integration (Section \ref{sec:diff_training}).

\subsection{Architecture Design}
\label{sec:arch}
The dynamics of an advection-diffusion-reaction system can be described by the following form of PDE:
\begin{equation}
\label{eqn:energy_eqn}
    \frac{\partial{\mathbf{x}}}{\partial{t}} = k\Delta \mathbf{x} -\mathbf{u} \cdot \nabla{\mathbf{x}} +  \mathbf{R_{x}(\mathbf{x},\mathbf{u}, \mathbf{c})} \, ,\\   
\end{equation}
where $\mathbf{x}(\mathbf{r}, t)$ is the physical variable of interest at position $\mathbf{r}\in\mathbb{R}^d$ and time $t$, and the operators $\nabla$ and $\Delta := \nabla^2$ denote the gradient and the Laplacian, respectively. $\mathbf{u} \cdot \nabla{\mathbf{x}}$ is the advection term and $k\Delta\mathbf{x}$ is the diffusion term with the diffusivity coefficient $k$. $\mathbf{R_{x}} (\mathbf{x}, \mathbf{u}, \mathbf{c})$ represents the generation or consumption sources for the state fields, with $\mathbf{c}$ denoting a set of constant parameters. 

Eq. \ref{eqn:energy_eqn} is a generic form of PDE that covers a variety of different physical phenomena. For example, by setting the reactive term $\mathbf{R}_\mathbf{x}$ to zero, one may obtain the Burgers' equation, or by setting $\mathbf{u} \cdot \nabla{\mathbf{x}}=0$ and $\mathbf{R_{x}} (\mathbf{x}, \mathbf{u}, \mathbf{c}) = 0$, one may obtain the simple diffusion equation. Additionally, beyond physics, Eq. \ref{eqn:energy_eqn} is also equivalent to the Fokker–Planck equation in stochastic process modeling, or closely related to the Black–Scholes equation in financial dynamics, which appears in various machine learning literature.

In addition to Eq. \ref{eqn:energy_eqn}, the velocity field $\mathbf{u}$ may also be governed by the momentum equation in many physical systems: 
\begin{equation}
\label{eqn:momentum_eqn}
    \frac{\partial{\mathbf{u}}}{\partial{t}} = - \mathbf{u} \cdot \nabla{\mathbf{u}} + \mathbf{R_{u}(\mathbf{x},\mathbf{u}, \mathbf{c})},\\ 
\end{equation}
where $\mathbf{R_{u}(\mathbf{x},\mathbf{u}, \mathbf{c})}$ term encapsulates the forces applied to a unit volume element.

The system of PDEs comprised of Eqs. \ref{eqn:energy_eqn} \& \ref{eqn:momentum_eqn} is usually accompanied by the following initial conditions: 
\begin{equation}
\label{eqn:intital_condition}
    \arraycolsep=1.4pt\def\arraystretch{1.5}
    \left\{
    \begin{array}{ccl}
        \mathbf{u}(t = 0) &= &\mathbf{u}_0 \\
        \mathbf{x}(t = 0) &= &\mathbf{x}_0,
    \end{array}
  \right.
\end{equation}
and this initial value problem can be solved numerically using the forward update scheme:
\begin{equation}
\label{eqn:solution}
    \arraycolsep=1.4pt\def\arraystretch{1.5}
    \left\{
    \begin{array}{ccl}
        \mathbf{u}_{k+1} &= &\mathbf{u}_k + \int_{t_k}^{t_{k+1}}{F_{\mathbf{u}}(\mathbf{u},\nabla \mathbf{u}, \mathbf{x},\nabla\mathbf{x}, \mathbf{c})dt} \\
        \mathbf{x}_{k+1} &= &\mathbf{x}_k + \int_{t_k}^{t_{k+1}}{F_{\mathbf{x}}(\mathbf{u}, \mathbf{x},\nabla \mathbf{x}, \mathbf{c})dt} , \\
    \end{array}
    \right. \\
\end{equation}
where: 
\begin{equation}
\label{eqn:differentiator}
    \arraycolsep=0.8pt\def\arraystretch{1.5}
    \left\{
    \begin{array}{ccl}
        F_{\mathbf{u}} &= &- \mathbf{u}_{k} \cdot \nabla{\mathbf{u}}_{k} + \mathbf{R_{u}}(\mathbf{x}_{k},\mathbf{u}_{k}, \mathbf{c}) \\
        F_{\mathbf{x}} &= &- \mathbf{u}_{k} \cdot \nabla{\mathbf{x}_{k}} + k \Delta \mathbf{x}_k + \mathbf{R_{x}}(\mathbf{x}_{k},\mathbf{u}_{k}, \mathbf{c}). \\
    \end{array}
  \right.
\end{equation}

Here, to minimize the notational burden, we hereinafter denote the time integration $\int_{t_k}^{t_{k+1}} F_\mathbf{u}dt$ and $\int_{t_k}^{t_{k+1}} F_\mathbf{x}dt$ as shorthand integral operators $\Psi_\mathbf{u}$ and $\Psi_\mathbf{x}$, respectively.


Illustrated in Figure \ref{fig:architecture} is a realization of this formulation in a recurrent neural network. As shown in the purple box of the figure, PARCv2 approximates the governing PDEs, namely $F_\mathbf{x}$ and $F_\mathbf{u}$ in Eq. \ref{eqn:differentiator}, using CNN layers. The differentiator module (the purple box) takes the physical state $\mathbf{x}(\mathbf{r}, t)$ at time $t$ as input alongside the velocity field $\mathbf{u}(\mathbf{r}, t)$. It then predicts the future state $\mathbf{x}(\mathbf{r}, t+\Delta t)$ and the future velocity field $\mathbf{u}(\mathbf{r}, t+\Delta t)$ after the time interval $\Delta t$. Here, the differential operators $\nabla$ and $\Delta$ in the advection and diffusion terms are approximated via the central finite difference scheme. The reaction functions $\mathbf{R_{x}}$ and $\mathbf{R_{u}}$ are modeled using separate CNN layers. 
Subsequently, the integrator (the orange box in Figure \ref{fig:architecture}) integrates the time derivatives $\dot{\mathbf{x}}$ and $\dot{\mathbf{u}}$ outputted by the differentiator to update the fields $\mathbf{x}$ and $\mathbf{u}$. The details on the integrator are presented in the following section.


\subsection{Hybrid Integration}
\label{sec:data_driven_int}
For physical systems with unsteady and fast transient characteristics, simulations can quickly deviate from the true dynamic trajectories as numerical and modeling errors accumulate during time integration.
Traditional numerical methods for initial value problems such as the Runge-Kutta methods or the Adams methods
take multiple steps in time to correct the error accumulation heuristically.
However, unsteady and fast transient problems require a higher-order correction with a small time step, making it less tractable for our purpose of rapidly estimating the dynamics. The original PARC \cite{nguyen2023sciadv} and other works in literature (\textit{e.g.}, \citet{shen2020deep}) explored the use of a deep neural network-based quadrature for time integration. However, as we will demonstrate in Sections \ref{sec:exp_ns} \& \ref{sec:exp_em}, these purely data-driven integrators fail to predict advection features with sharp boundaries precisely.

To address such limitations, we propose a hybrid approach, in which the low-order truncation errors are corrected using a conventional numerical integration, whereas higher-order errors are learned from data and compensated via CNN layers:
\begin{equation}
\label{eqn:data_driven_integration}
    \arraycolsep=1.4pt\def\arraystretch{1.5}
    \left\{
    \begin{array}{ccl}
        \mathbf{u}_{k+1} &= &\mathbf{u}_k + \Psi_{u} + S_{u}(\mathbf{u}_{k},F_{u}|\phi_{u}) \\
        
        \mathbf{x}_{k+1} &= &\mathbf{x}_k + \Psi_{x} + S_{x} (\mathbf{x}_{k},F_{x}| \phi_{x}) . \\
    \end{array}
  \right.
\end{equation}
Here, $\Psi_{u}$ and $\Psi_{x}$ are the integrals computed by a conventional numerical method, whereas $S_{u}$ and $S_{x}$ represent higher-order terms approximated by convolutional neural networks with network parameters $\phi_{u}$ and $\phi_{x}$.
Here, the choice of numerical integration method for $\Psi_{u}$ and $\Psi_{x}$ depends on the tradeoff between the complexity of the problem and computational speed. For instance, we found that the second-order forward Euler scheme (Heun's method) worked reasonably well for Burgers' (Section \ref{sec:exp_burgers}) and Navier-Stokes equations (Section \ref{sec:exp_ns}), while for the energetic materials problem (Section \ref{sec:exp_em}), the fourth-order Runge-Kutta (RK4) method was necessary.

\subsection{Training}
\label{sec:diff_training}

In PARCv2, the differentiator and the integrator are trained separately. We empirically found that the two-stage training yielded more accurate prediction results than end-to-end training. While this empirical observation is partially supported by the previous work of \citet{Poli2020} on the hypersolvers for Neural ODE, we do not have a concrete explanation in this regard and hence this tendency awaits future investigation.

In this two-stage training process, the differentiator is trained without the data-driven higher-order error terms $S_{u}$ and $S_{x}$, but only with the numerical integrals $\Psi_{u}$ and $\Psi_{x}$ in the loss function:
\begin{equation}
\begin{split}
    \mathcal{L}(\theta_\mathbf{u}, \theta_\mathbf{x}) := \sum_{t_k} {|| \hat{\mathbf{u}}_{k+1} - \hat{\mathbf{u}}_{k} - \Psi_{\mathbf{u}}\left[ F_u(\hat{\mathbf{u}}_{k}|\theta_\mathbf{u})\right] ||}_1 \\ + \sum_{t_k} {|| \hat{\mathbf{x}}_{k+1} - \hat{\mathbf{x}}_{k} - \Psi_{\mathbf{x}}\left[F_x(\hat{\mathbf{x}}_{k}|\theta_\mathbf{x})\right] ||}_1.
\end{split}
\label{eqn:diff_train}
\end{equation}
Here, the upper hat denotes the ground truth. Gradient-based optimization is employed for training, in which the gradients are computed by differentiating through the integration operations of the forward pass.  

Once the differentiator is trained, the weights are frozen. Then the data-driven integrals $S_{u}$ and $S_{x}$ are trained with the following training loss:
\begin{equation}
\begin{split}
\mathcal{L}(\phi_\mathbf{u}, \phi_\mathbf{x}) := \sum_{t_k} {|| \hat{\mathbf{u}}_{k+1} - \hat{\mathbf{u}}_{k} - \Psi_{\mathbf{u}}} -  S_{u}(\mathbf{u}_{k},F_{u}|\phi_{u}) ||_1 \\ +   \sum_{t_k} || \hat{\mathbf{x}}_{k+1} - \hat{\mathbf{x}}_{k} - \Psi_{\mathbf{x}} -  
S_{x}(\mathbf{x}_{k},F_{x}|\phi_{x})||_1   
\end{split}
\label{eqn:integration_training}
\end{equation}
where $\hat{u}_{k+1}$ and $\hat{x}_{k+1}$ are ground truth values of velocity and state variables at time $t_{k+1}$ and $\hat{u}_{k}$ and $\hat{x}_{k}$ are ground truth values at time $t_k$.

\section{Results}
\label{sec:results}
\begin{table*}[t]
\caption{Prediction accuracy of PARCv2 and the other baseline models. The derivation of the metrics reported in the Table is given in the Appendix (Section \ref{sec:metrics})}
\label{tab:pred_acc}
\vskip 0.15in
\begin{center}
\begin{small}
\begin{sc}
\begin{tabular}{lcccc}
\toprule
Model           & Burgers          & Navier-Stokes           & \multicolumn{2}{c} {Energetic Materials} \\
                & $RMSE_u (cm/s) $     &  $RMSE_u (m/s)$            & $RMSE_T (K) $         & $RMSE_P (GPa)$ \\

\midrule
PARC (numerical int.)  & 0.0074                & 0.2336             &  249.99 &    1.491\\
PARC (data-driven int.)  & 0.0236                & 0.3089             &  306.99 &    4.111\\
FNO             & 0.0289               & 0.2147              &   248.39 & 2.685\\
PhyCRNet        & 0.0588                & 0.2094              & - & - \\
PIFNO          & 0.0338                & 0.2307            &  - & - \\  
PARCv2 (This Study)   & 0.0129                & 0.0727             &  229.52& 1.634\\

\bottomrule
\end{tabular}
\end{sc}
\end{small}
\end{center}
\vskip -0.1in
\end{table*}

\begin{table*}[t]
\caption{Solution quality of PARCv2 and the other baseline models.The process of computing the metrics introduced in the table is given in the Appendix (Section \ref{sec:metrics})}
\label{tab:sol_qual}
\vskip 0.15in
\begin{center}
\begin{small}
\begin{sc}
\begin{tabular}{lccccccc}
\toprule
Model     & Burgers & \multicolumn{2}{c}{Navier-Stokes} & \multicolumn{4}{c} {Energetic Materials} \\
          & $||\mathbf{f_u}||$   & $||\mathbf{f_u}||$ &  $\varepsilon_{div}$ & $\varepsilon_{T^{hs}} $ & $\varepsilon_{A^{hs}} $ & $\varepsilon_{\dot{T}^{hs}}$& $\varepsilon_{\dot{A}^{hs}} $ \\
          & $ (cm/s^2)$            & $ (m/s^2)$ &  $ (1/s)$  & $(K)$ & $(\mu m^2)$ & $(K/ns)$& $ (\mu m^2/ns)$ \\
\midrule
DNS       & 0.1241 & 2.7041 & 0.0398             & - & - & - & - \\

PARC (numerical int.) & 0.1262 & 2.8568 & 0.4465 & 409.58 & 0.0253 & 269.09 & 0.0248 \\
PARC (data-driven int.)  & 0.1176 & 6.2244 & 4.3241 & 972.38 & 0.0728 & 839.64 &0.0681 \\

FNO    & 0.1537 & 2.5351 & 0.9170& 622.60 & 0.0431 & 425.91 & 0.0527 \\
PhyCRNet    & 0.0560 & 2.4904 & 0.0565 & - & - & - & - \\
PIFNO    & 0.1058 & 0.8552 &  0.0768 & - & - & - & - \\  
PARCv2 (This Study)  & 0.1292 & 3.7095 & 0.3738 & 149.27 & 0.0060 & 228.98 & 0.0094 \\

\bottomrule
\end{tabular}
\end{sc}
\end{small}
\end{center}
\vskip -0.1in
\end{table*}

Here, we evaluate PARCv2's performance on different physics benchmark problems, namely the Burgers equation, cylindrical flow (Navier-Stokes equations), and pore-collapse in energetic materials. The Burgers' equation (Section \ref{sec:exp_burgers}) represents advection-diffusion systems with weak advection dominance, while the cylindrical flow problem represents systems with unsteady dynamics and strong advection dominance. The energetic materials problem serves as an archetype of fast transient systems with strong sharp gradients and rapidly deforming boundaries. Details on these benchmark problems can be found in the Appendix.

Baseline methods for comparison include the original PARC \citep{nguyen2023sciadv}, as well as state-of-the-art PADL methods including PhyCRNet \cite{ren2022phycrnet}, Fourier Neural Operator (FNO)~\citep{li2021fourier}, and FNO with a physics-informed loss (PIFNO). PARC was implemented in two different ways, one with purely data-driven integration (original PARC) and one with a traditional numerical solver (forward Euler), for the ablation study on the effect of different integration methods. For all benchmark models, we mostly adopted the implementation of the original authors, except for minor modifications made to make the algorithm compatible with the problem settings. All details are described in the Appendix.

\subsection{Burgers' Equation}
\label{sec:exp_burgers}

\begin{figure}[tb!]
     \centering
     \includegraphics[width=0.48\textwidth]{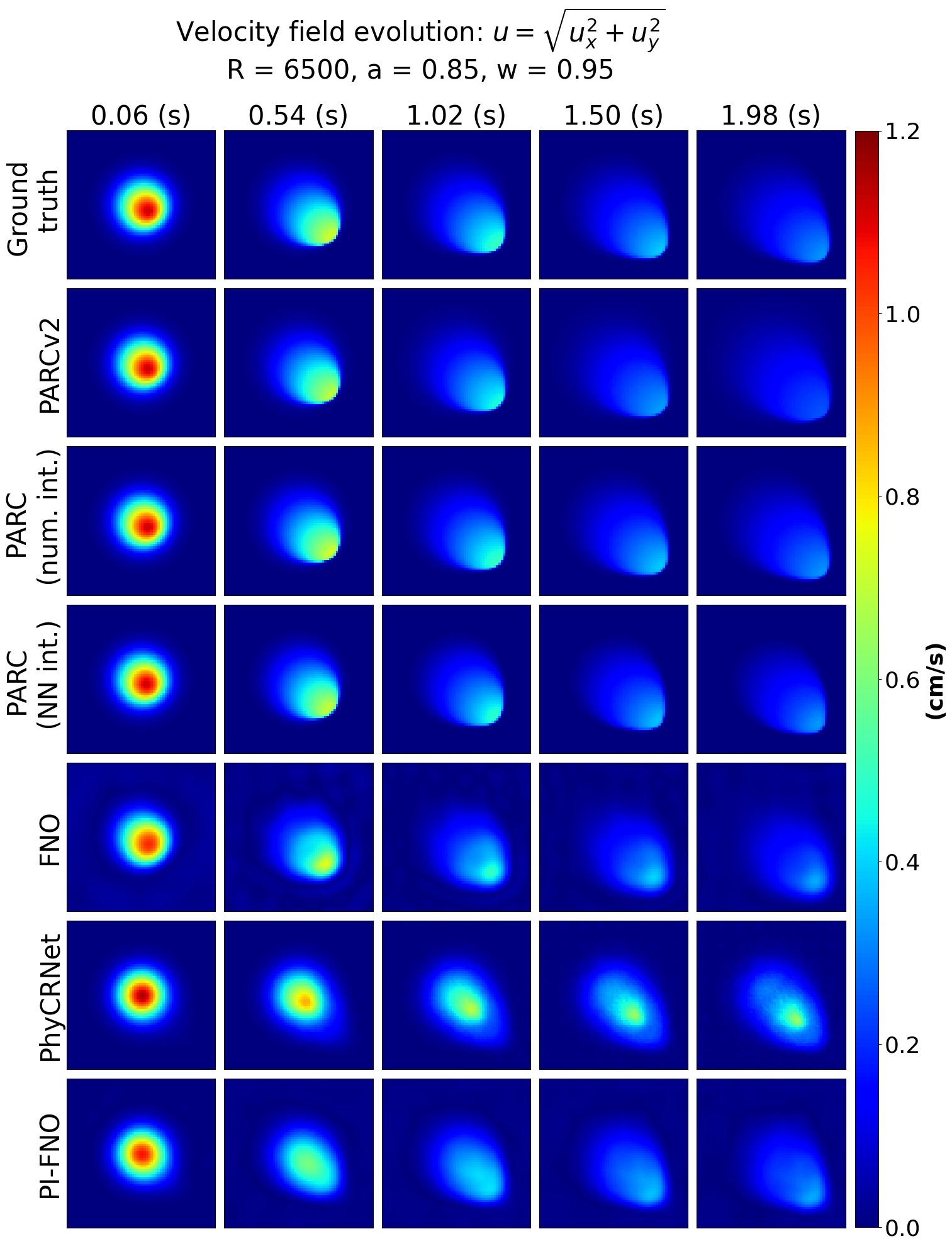}
\caption{Validation on the Burgers' equation.}
\label{fig:burgers}
\end{figure}

\begin{figure}[tb!]
     \centering
     \includegraphics[width=0.45\textwidth]{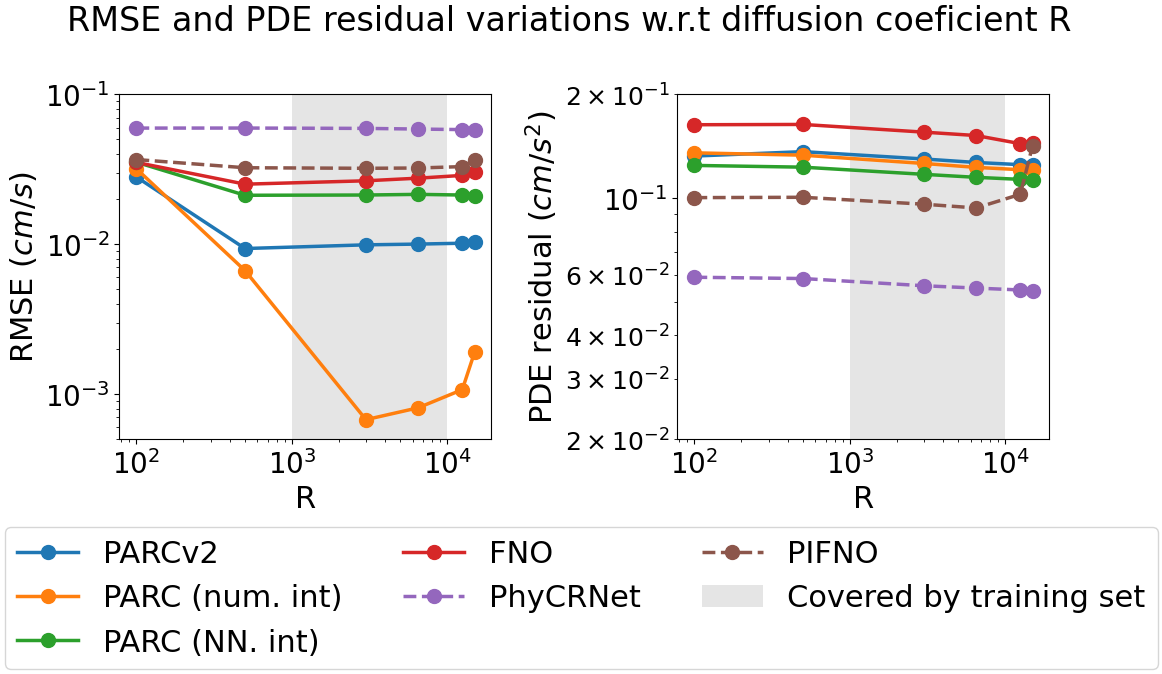}
\caption{Prediction accuracy (\textit{left}) and solution quality (\textit{right}) on the Burgers' equation cases with respect to varying diffusion coefficients $R$.}
\label{fig:burgers_mse_pde_vs_R}
\end{figure}

We first tested the capability of PARCv2 for solving Burgers' equation describing the formation and decay of a shock wave. The top row of Figure \ref{fig:burgers} displays the ground truth phenomenon simulated by a direct numerical solver (DNS). Notice the formation of a sharp front at the bottom-right corner and the decay of the velocity on the tail end (top-left). 
From the visual inspection of Figure \ref{fig:burgers}, both PARC and PARCv2 appear to predict the formation of the sharp front and the decay on the tail end, while other models tend to exhibit blurred and smeared fronts. 
This observation can further be confirmed by the quantitative results reported in the first column of Table \ref{tab:pred_acc}, in which both PARC and PARCv2 yielded root mean squared errors (RMSEs) of the predicted velocity field of around 0.010 cm/s, which is about half of the RMSEs of the other models.
Notably, employing the pure data-driven integration in the case of PARC yields less accurate predictions compared to the use of numerical integration with the same differentiator architecture and the hybrid integration of PARCv2 (first column of Table \ref{tab:pred_acc}).



In terms of solution quality, measured by the residual of Burgers' equation from the predicted fields (Table \ref{tab:sol_qual}), both PARC and PARCv2 demonstrated comparable results to other models. While PhyCRNet and PIFNO exhibited the lowest residuals, likely due to explicit minimization of the PDE residual through physics-informed training loss, the solution quality of PARC and PARCv2 was not significantly inferior, especially considering the ground truth data (DNS) already had a similar PDE residual of 0.1241 cm/s$^2$. The observed high solution quality and low prediction accuracy of PhyCRNet and PIFNO may result from over-regularization due to incorporating the PDE residual term in the training loss. Indeed, this concern has been previously documented in the literature, and the validation results presented here reaffirm its significance.





Additionally, Figure \ref{fig:burgers_mse_pde_vs_R} depicts trends in RMSE and PDE residuals across different diffusion coefficients $R$. The shaded area represents the range of $R$ covered by the training set. All models appear to be generalizable over different values of $R$, with an observation that PARC with a numerical integrator quickly deteriorated in its prediction accuracy (RMSE). Interestingly, the solution quality (PDE residual) stayed more or less the same. We suspect that this could be attributed to overfitting in PARC, as its training lacks the physics-informed loss term as a regularizer. However, due to the physics-aware architecture, PARC still ensures adherence to physical laws that are governed by the Burgers' equation.

Finally, it is worthwhile to note that despite choosing a relatively large discrete time step, which may appear to contravene the Courant-Friedrichs-Lewy (CFL) condition of conventional explicit solvers, the stability of the solutions by PARCv2 remains unaffected by this large step size selection. While lacking more solid evidence, we speculate that errors stemming from our discretization choices have been mitigated by both the differentiator and integrator of PARCv2, owing to their robust modeling capabilities.


\subsection{Navier-Stokes Equations} 
\label{sec:exp_ns}

\begin{figure}[tb!]
     \centering
     \includegraphics[width=0.485\textwidth]{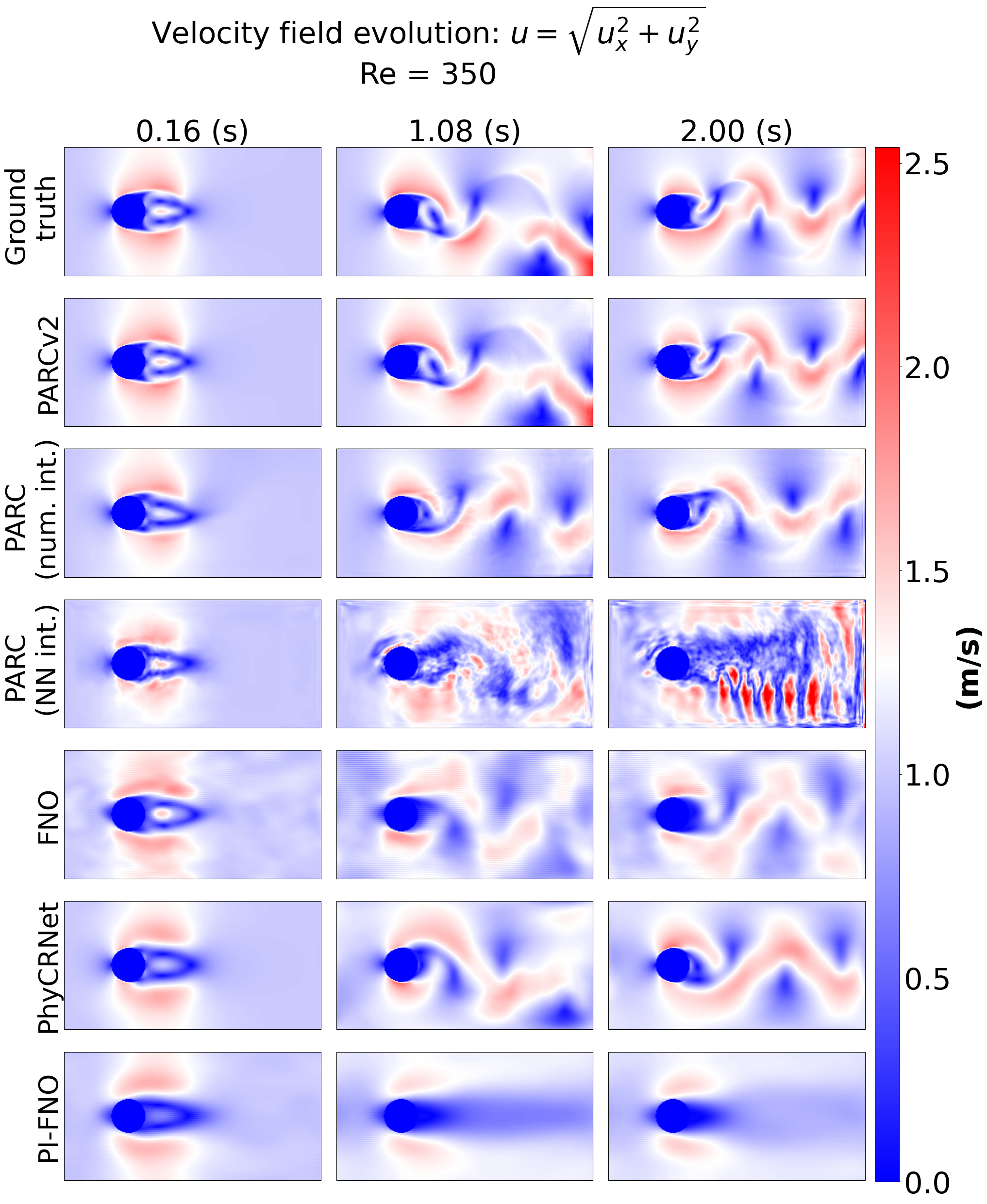}
\caption{Validation on the Navier-Stokes equation (cylindrical flow).}
\label{fig:ns}
\end{figure}

Another popularly used benchmark problem is the cylindrical flow, expressed with the Navier-Stokes equations. Compared to the Burgers' equation cases, this problem introduces additional complexity with the phenomenon of vortex shedding, resulting in unsteady flow fields with intricate and deformable boundaries.

Figure \ref{fig:ns} displays the fluid velocity fields predicted by the models. Visually, all methods appear to capture the oscillatory vorticity patterns to a certain extent. However, the oscillatory phase of the vorticity patterns was most accurate in PARCv2, highlighting its stable long-time prediction capabilities. This observation is further supported by Table \ref{tab:pred_acc}, where the RMSE of PARCv2 was 0.0727 m/s---roughly 1/4 of the values obtained by the other methods. 

Interestingly, the original PARC, only with a purely data-driven integrator (fourth row in Figure \ref{fig:ns}), exhibited highly noisy predictions. This may be indicative of the rapid accumulation of network prediction errors in unsteady flows with sharp features. 
It is also worthwhile to compare PARCv2 (second row in Figure \ref{fig:ns}) with the original PARC with the numerical integrator (third row in the same figure). Compared to the purely data-driven integration case, the original PARC exhibited a more stable prediction with the numerical integrator. However, the original PARC still appears to be off from the ground truth in terms of the oscillatory phase of the vorticity pattern, while PARCv2 aligns well with the true phase. This result highlights the importance of the data-driven higher-order correction in the integrator. Furthermore, not only the phase but also the geometric pattern slightly deviates from the ground truth in the case of the original PARC, in contrast to the PARCv2 result where the geometric patterns are faithfully reproduced. This is indicative of the role of the spatial gradient terms newly introduced in the PARCv2 architecture.



The solution quality of the PARC model family, however, lags notably behind the other physics-informed benchmark models (i.e., PhyCRNet and PIFNO). Particularly, PARCv2 tends to deviate noticeably from the divergence-free condition ($\varepsilon_{div}$), affecting its compliance with the incompressibility of the flow. In other words, although PARCv2 may visually produce more accurate vorticity patterns, it significantly violates the incompressibility constraint. On the contrary, the physics-informed benchmarks, while may look visually less accurate, at least can comply well with the physical constraints. Arguably, such a contrast between PARCv2 and physics-informed models appears to stem from the optimization problem during training. That is, gradients from the physics-informed loss (PDE residual) and data loss (RMSE) often conflict, hindering model convergence. In contrast, PARCv2 does not face such conflicts but may not fully adhere to physics constraints upon convergence. Hence, there is an opportunity where physics-informed loss terms could be supplemented to PARCv2 training after the initial convergence is made. However, determining the optimal trade-off between the two aspects requires further research beyond the scope of this paper.


\subsection{Energy Localization in Energetic Materials}
\label{sec:exp_em}

\begin{figure}[tb!]
\centering
\begin{subfigure}{0.475\textwidth}
     \centering
     \includegraphics[width=\textwidth]{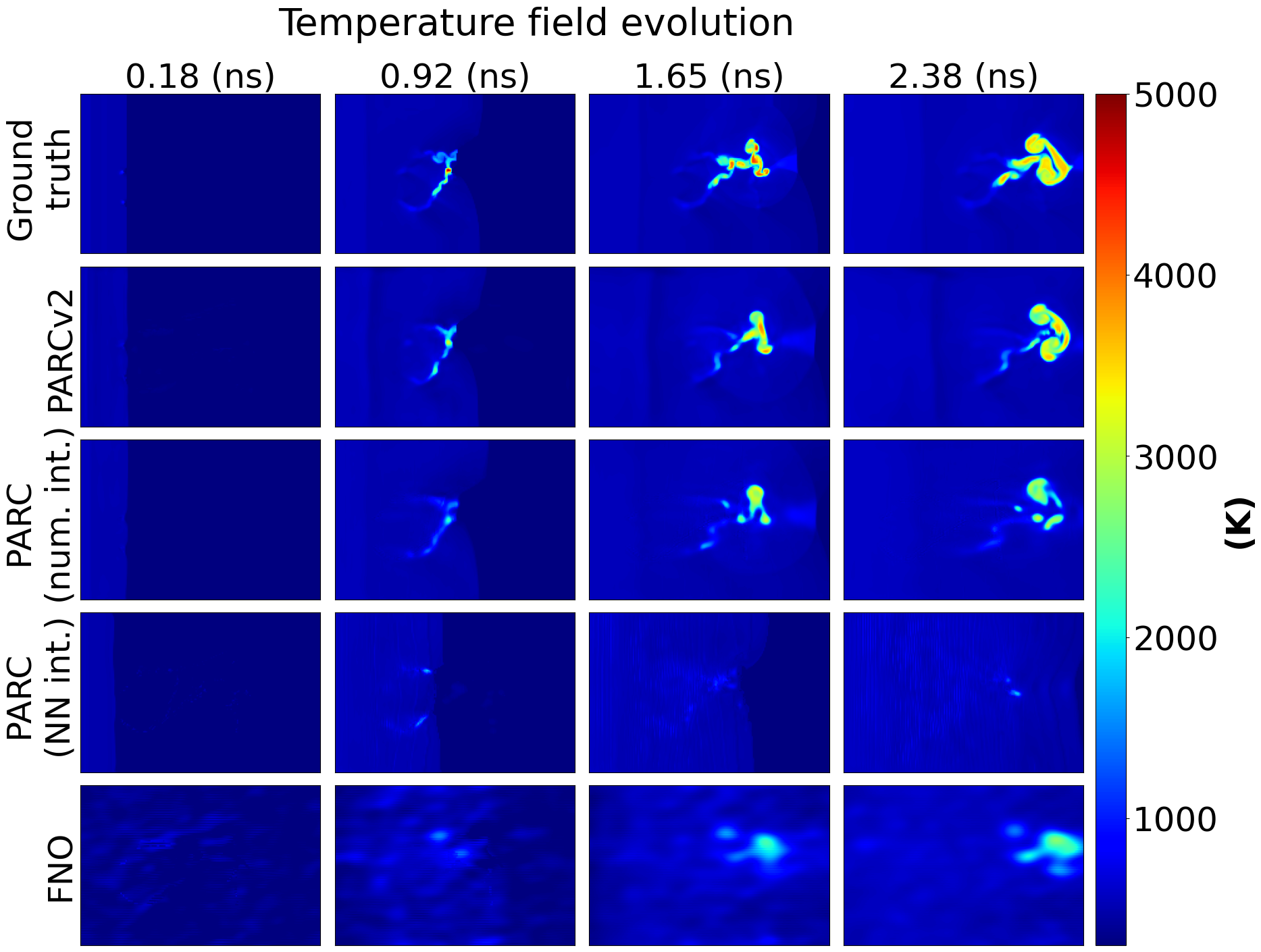}
     \label{fig:temp}
\end{subfigure}
\hfill
\begin{subfigure}{0.48\textwidth}
     \centering
     \includegraphics[width=\textwidth]{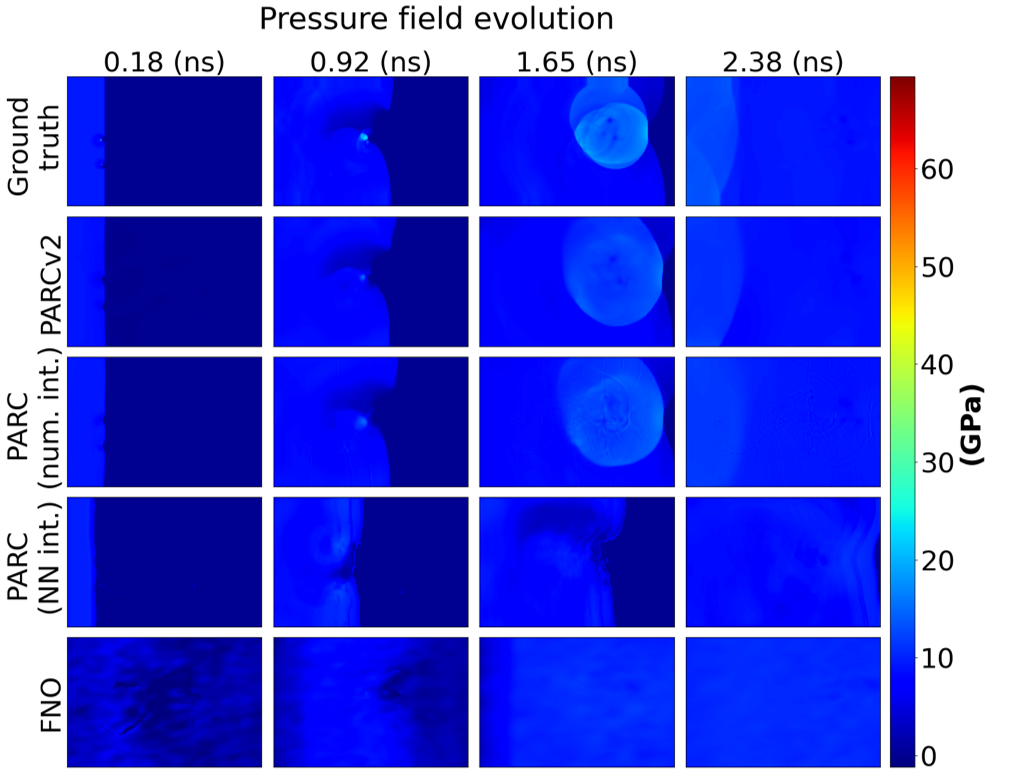}
     \label{fig:press}
\end{subfigure}
\caption{Prediction of energy localization in energetic materials. }
\label{fig:em}
\end{figure}

Energetic materials, such as explosives and propellants, exhibit unsteady and fast transient thermomechanics with abrupt changes in temperature and pressure fields during their initiation process. Furthermore, physics-based simulation of energetic materials may take days on high-performance clusters. These issues pose an interesting machine-learning challenge, in which unsteady and fast transient field evolutions must be learned from limited data. The presence of sharp state field gradients, including traveling shocks and evolving boundaries, further intensifies the complexity. Moreover, the lack of an explicit equation for the thermodynamics at the mesoscale is another obstacle that one needs to overcome.


In this problem, while the details are deferred to Appendix (Section \ref{sec:em_setup}), we are interested in predicting the evolution of temperature $T$ and pressure $P$ fields, given the material condition $\mu$. Explicitly defined governing equations describing the changes in temperature and pressure as functions of $T$, $P$, and $\mu$ are unavailable. Therefore, one must depend solely on data to learn the underlying physics. It is crucial to highlight that the lack of a governing equation makes it impossible to define the physics-informed loss term for learning-bias models like PhyCRNet and PIFNO. Without a PDE, PIFNO becomes identical to the regular FNO and PhyCRNet is essentially a convolutional long short-term memory (LSTM) model \citep{ConvLSTM-Shi} that is physics na\"ive and there is no need for a separate discussion. Hence, these models were excluded from the experiment.

Figure \ref{fig:em} illustrates the temperature and pressure field evolution predictions made by PARCv2, PARC (with the numerical integrator and the data-driven integrator), and FNO. As shown in the figure, PARCv2 predictions reproduced the most amount of details in terms of the position and shape of temperature ``hotspots'' (localized high-temperature regions). From the comparison with the original PARC results, it is observable the efficacy of the newly introduced spatial gradient terms in the PARCv2 architecture. Furthermore, with hybrid time integration, PARCv2 effectively reproduces the shape and boundary of the hotspot in the temperature field evolution.

The quantitative results in Table \ref{tab:pred_acc} further supported this observation. For the temperature RMSE, there was no significant difference between PARCv2 ($\sim$230 K) and the other two methods ($\sim$250 K). In the case of the pressure RMSE, the difference between PARC with numerical integration ($\sim$1.5 GPa) and FNO ($\sim$2.7 GPa) was more prominent. More noteworthy are the solution quality metrics reported in Table \ref{tab:sol_qual}. In this case,
we used the average temperature and area of hotspots ($T^{hs}$, $A^{hs}$) and their growth rates ($\dot{T}^{hs}$, $\dot{A}^{hs}$) as the measures of solution quality, following the standard practice in the energetic materials community \citep{Nguyen_single_void, Seshadri}. For these metrics, the performance difference between PARCv2 and the other two benchmarks was more prominent.

\section{Discussion}
The enhanced performance found in PARCv2 compared to other baselines across all benchmark problems may be explained theoretically as follows. 

Borrowing the views of mathematical physics and differential geometry, an evolving field of a physical quantity can be treated as a \textit{trajectory}, or a \textit{curve}, in the Galilean spacetime. A submanifold embedded in the Galilean spacetime would describe some physical rules and constraints. That is, if the physical quantity is subjected to such rules and constraints, the corresponding trajectory or curve must lie on the manifold. Additionally, when there are multiple observations of such evolving fields under different conditions, these observations create a visual of a bundle of curves all of which are lying on the dynamics manifold.

In this setting, PARCv2 can be viewed as learning the topological structures of the dynamics manifold, in contrast to the learning bias approaches that enforce physics via geometric constraints in spacetime. Particularly, PARCv2 attempts to learn the approximation of the tangent spaces on the dynamics manifold (tangent bundle), as well as a ``cone'' in each tangent space describing where the dynamics may head at a given state. On the contrary, other physics-informed models aim to fit the manifold itself directly, by evaluating the manifold positions across different sample points, or what the literature refers to as ``collocation points.''

Given this theoretical view, we argue that the problem of learning the tangent bundle and the tangent cones therein in the case of PARCv2 is a simpler optimization problem, compared to fitting the manifold directly such as in other baselines. The empirical evidence on multiple benchmark problems supports this argument, as discussed in Section \ref{sec:results} about the ``tension'' between prediction error (RMSE) and solution quality (PDE residuals). When the dynamics manifold is relatively smooth, previous PIML methods may approximate the manifold geometry effectively. However, in the presence of strong nonlinearity and limited data, fitting the solution manifold globally becomes extremely challenging. PARCv2 addresses this by imposing hard constraints on the topology of the dynamics manifold and conducting local physics-informed regularization, allowing for better learning capability in the presence of strong nonlinearity. This approach narrows the search space of PARCv2 learning, resulting in enhanced performance compared to previous PIML methods in handling strong nonlinearity.

\section{Conclusion}
This study extended PARC to a broader range of spatiotemporal dynamics phenomena. Through the incorporation of new advection and reaction terms, as well as the introduction of a hybrid integration solver, PARCv2 demonstrated improved predictive performance compared to the original PARC and previous learning bias approaches on various benchmark physics problems. The validation experiments also revealed that 
the PARCv2 differentiator-integrator architecture can produce results that comply with underlying physics laws, with a comparable solution quality to the other learning-bias approaches, despite no explicit PDE residual terms during training. Finally, PARCv2 was shown to be capable of discovering unknown physics in the absence of explicitly defined governing equations. 

These advantages underscore the significance and potential of the inductive bias research direction, positioning it as a promising complement to current mainstream learning bias approaches. It is, in fact, important to note that PARCv2 is not positioned as an exclusive alternative to learning-bias methods, as the incorporation of physics-informed loss during PARC training is still possible. Future research directions may explore the synergistic leverage of soft constraints imposed by the physics-informed loss and the inductive bias provided by the PARC differentiator-integrator architecture. Enhancing the computational efficiency of PARCv2 is also another opportunity for improvement. From our experiments, PARCv2 is slightly heavier than other PIML models, while still significantly less expensive than direct numerical simulations. To this end, potential valuable research directions would include reduced-order modeling (\textit{i.e.}, dimension reduction or learning the latent space dynamics). 

Beyond applied physics research, we also envision the broader use of PARCv2 in computer vision and machine learning contexts. 
The flexibility of not requiring the explicitly defined governing quations enables the application of PARCv2 to various problems in computer vision and machine learning framed as physics or stochastic processes, such as diffusion modeling for image generation or neural ordinary differential equations.

Finally, considering the potential societal impacts of PARCv2 is also crucial. For one, interpretability and visualization of PARCv2 is what we deem as the most urgent issue, as it can shed light on how reliable and trustworthy the predictions are. Furthermore, the ability to rapidly explore complex physical phenomena may also be a double-edged sword, as the characterization of sensitive materials, for example, may result in various consequences. Therefore, PARCv2 should be used carefully to avoid any unwanted negative societal impact.

\section*{Impact Statement}
This paper presents work whose goal is to advance the field of Machine Learning. There are many potential societal consequences of our work, none of which we feel must be specifically highlighted here.

\section*{Acknowledgements}
We appreciate the anonymous reviewers for their comments on improving this work.
This work was supported by the Army Research Office (ARO) Energetics Basic Research Center (EBRC) program under Grant No.~W911NF-22-2-0164, the National Science Foundation under Grant No. DMREF-2203580, and the Department of Energy under contract No. DE-AC52-07NA27344.

\bibliographystyle{unsrtnat}
\bibliography{main.bib}  

\newpage
\appendix
\onecolumn
\section{Appendix}
\subsection{Burgers' Equation}
\label{sec:burgers_setup}
\textit{Burgers' equation} is a convection-diffusion equation that occurs in a variety of applications such as fluid mechanics, acoustics, plasma physics, and traffic flows. With the presence of the viscous (diffusion) term, the shock discontinuities are smoothed out and one can expect to obtain a well-behaved and smooth solution which will make the learning task less challenging for PARCv2 and other models. Therefore, learning to solve Burgers' equation can be considered the most straightforward compared to the other two validation studies and was used as the first experiment for our validation.

The two-dimensional Burgers' equation is given as:
\begin{equation}
\label{eqn:2d_burgers}
\begin{split}
\frac{\partial u}{\partial t} &= -u\cdot \nabla u + \frac{1}{R}\Delta  u\\
\frac{\partial v}{\partial t} &= -v\cdot \nabla v + \frac{1}{R}\Delta  v
\end{split}
\end{equation}
where $\mathbf{u}(\mathbf{r}, t):=[u(\mathbf{r}, t), v(\mathbf{r}, t)]^\top$ is the velocity of fluid at position $\mathbf{r}$ and time $t$ and $R$ is the Reynolds number.

In the experiment presented in this paper, we assume that the fluid velocity is initially distributed according to the following initial condition:
\begin{equation}
\label{eqn:2d_burgers_ic}
     u(\mathbf{r},t)|_{t=0} = v(\mathbf{r},t)|_{t=0} = a \exp(-||\mathbf{r}||^2_2/w)
\end{equation}

Additionally, for all experiments, we consider the domain with a spatial dimension of $6 \text{ cm}\times6 \text{ cm}$, resolved on a $64 \times 64$ grid ($1 \text{ pixel} = 0.09375 \text{ cm}$). The total simulation time is 2 seconds, uniformly divided into $100$ discrete time steps of length 0.02 seconds. The zero-velocity boundary condition is also applied to the spatial domain, such that $\mathbf{x}_0(\mathbf{r},t) = 0, \forall \mathbf{r} \in \partial \Omega$.

The fluid parameters, \textit{i.e.}, the Reynolds number $R$ and the initial velocity distribution $a$ and $w$ are varied according to the values listed in Table \ref{tab:burgers_setting}. Note that the fluid parameters in the train-test split are designed to test the models' abilities to not only interpolate within the training distributions but also extrapolate/generalize to conditions outside the training distributions.

\begin{table}[tbh!]
\caption{PDE constants for 2D Burgers' equation used for training and testing}
\label{tab:burgers_setting}
\vskip 0.15in
\begin{center}
\begin{small}
\begin{sc}
\begin{tabular}{lcc}
\toprule
                & Training & Testing \\
\midrule
 $R$ $(cm^2/s)$    & 1000, 2500, 5000, 7500, 10000& 100, 500, 3000, 6500, 12500, 15000 \\
$a$ $(cm/s)$      & 0.5, 0.6, 0.7, 0.8, 0.9 & 0.35, 0.40, 0.45, 0.55, 0.65, 0.75, 0.85, 0.95, 1.00\\
$w$ $(cm)$       & 0.7, 0.8, 0.9, 1.0 & 0.55, 0.6, 0.65, 0.75, 0.85, 0.95, 1.05\\
\bottomrule
\end{tabular}
\end{sc}
\end{small}
\end{center}
\vskip -0.1in
\end{table}

The training data was obtained by solving Eq.~\ref{eqn:2d_burgers} with numerical simulation using implicit backward Euler integration. Data was generated using the open-source code provided by \citet{fries2022lasdi}. The original numerical simulation data included 1,500 snapshots and was down-sampled to 100 snapshots to be used for training. The spatial resolution of the numerical simulation data was kept during the training. 

\subsection{Navier-Stokes Equation}
\label{sec:ns_setup}
Fluid flow around an obstacle has been a standard benchmark problem in fluid mechanics for decades, with laminar vortex shedding emerging as a particularly interesting and extensively studied phenomenon. Flow separation and vortex shedding occur due to high positive pressure gradients, resulting in flow fields characterized by relatively sharp features that evolve rapidly in space, especially at high Reynolds numbers. In comparison to the Burgers' equation case, the flow around an obstacle exhibits more advective characteristics, posing additional challenges for machine learning, as discussed earlier. Therefore, we selected this benchmark problem to assess and validate the modeling capability of PARCv2, building on its success with the Burgers' equation.

The \textit{Navier-Stokes equations} are a system of PDEs describing the mechanical behavior of fluids. The Navier-Stokes equations for the motion of an incompressible viscous fluid with constant density consist of a continuity equation for conservation of mass and an equation for conservation of momentum:
\begin{equation}
\begin{split}
     \nabla \cdot \mathbf{u} = 0 \quad&\text{(continuity)}\\
     \frac{\partial \mathbf{u}}{\partial t} + (\mathbf{u}\cdot \nabla) \mathbf{u} = -\frac{1}{\rho} \nabla p + \frac{1}{Re}\Delta  \mathbf{u}  \quad&\text{(momentum)}
\end{split}
\end{equation}
Here, $\mathbf{u}(\mathbf{r},t)$ denotes the velocity vector at position $\mathbf{r}$ and time $t$, $p(\mathbf{r}, t)$ the pressure, and $\rho$ and $R$ the density and Reynolds number respectively.

Oftentimes, the magnitude of the Reynolds number $R$ has a significant influence on the stability of fluid. When $R$ is small, the flow is inert and sluggish, rendering a rather stable flow pattern. On the other hand, when $R$ is large, the flow becomes unstable and starts to render complex patterns.

\begin{figure}[tbh!]
     \centering
     \includegraphics[width=0.35\textwidth]{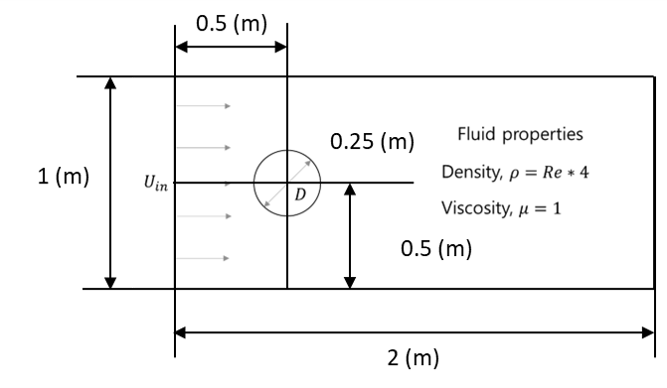}
\caption{Cylindrical flow benchmark.}
\label{fig:ns_appendix}
\end{figure}

In many physics-aware deep learning works in literature, a frequently used benchmark is how well the model can predict unstable fluid behavior with large Reynolds numbers. Similar to these works, we consider a cylindrical cross-section of 0.25 m diameter embedded in a 2 m $\times$ 1 m rectangular domain, as illustrated in Figure \ref{fig:ns_appendix}. The circular obstacle was positioned $0.5$ m far from the left and $0.5$ m from the top edge. We assume the initial velocity field to be uniformly distributed at the left side of the domain with initial velocity $u_0 = 1$ m/s. Here the fluid density, $\rho$, is derived as: $\rho = 4 \cdot R$ kg/m$^3$. Meanwhile, the kinematic viscosity, $\nu$, is derived from the dynamic viscosity $\mu$ as $\nu = \frac{\mu}{\rho}$.

For the train-test split, the Reynolds number was varied according to Table \ref{tab:ns_settings}. The training data was obtained using finite volume methods with ANSYS Fluent. The initial data was provided in the triangular mesh format and was rasterized to fit with the use of CNN using interpolation. The rasterized domain has the size of $128$ pixels $\times\, 256$ pixels with each pixel corresponding to a $7.8125\, \mu$m $\times 7.8125\, \mu$m area. 

\begin{table}[tbh!]
\caption{Fluid constants for 2D Navier-Stokes equation}
\label{tab:ns_settings}
\vskip 0.15in
\begin{center}
\begin{small}
\begin{sc}
\begin{tabular}{lcc}
\toprule
                & Training & Testing \\
\midrule
Dynamic viscosity  $\mu$ $(kg/(ms))$    & 1 & 1 \\
Inlet velocity $u_{0}$ $(m/s)$      & 1 & 1 \\ 
Reynold number $Re$        & 30,40,80,100,150,200,250,300, & 20,60,140,350,550,750,1000\\
                                & 400,450,500,600,650,700,800, & \\
                                & 850,900,950 & \\
\bottomrule
\end{tabular}
\end{sc}
\end{small}
\end{center}
\vskip -0.1in
\end{table}

\subsection{Energy Localization of Heterogeneous Energetic Materials}
\label{sec:em_setup}
Energetic materials (EM), such as explosives and propellants, exhibit intensive and transient thermomechanics. In particular, the localization of energy during the reaction dynamics of EMs results in high-temperature regions called ``hotspots.'' The formation and growth of hotspots under varying input shock loading conditions are important design considerations for EMs that determine the overall performance and safety of EMs.  
Here, the morphology of EM microstructures is known to play a crucial role in the formation and growth of hotspots. For this reason, the problem of predicting the hotspot formation and growth patterns from a microstructure image is an important problem for the safe design of EMs.

From the machine learning perspective, this problem represents a typical instance of real-world extreme physics, which can present significant challenges for physics-informed machine learning. Not only is there an absence of a well-defined governing equation to apply a physics-informed loss, but the intense and transient thermomechanics also result in strong and sharp spatiotemporal gradients of temperature and pressure field evolution, the two state variables of interest. Furthermore, the sharp gradient state fields in this problem may encompass traveling shocks, collapsing interfaces, and evolving internal/external boundaries that interact with the flow field. These characteristics pose a significant challenge for machine learning, making the training process more difficult to converge to the target dynamics. Given these challenges rendered by the nature of the problem, we have chosen this problem as our final experiment to validate the modeling capability of PARCv2 in the realm of extreme physics.

In this study, we employ a similar setup to the previous work of \citet{nguyen2023sciadv,nguyen2023pep}, in which we consider a reduced dynamical system that includes state variables $\mathbf{x}:=[T, P,\mu]$ and velocity variables $u:=[u_{x},u_{y}]$. In this system, $T$ represents the temperature, $P$ represents the pressure, and $\mu$ represents the microstructure density. It is important to note that the specific form of the governing PDEs for this reduced system is unknown and will be learned from data. However, based on our domain knowledge, the reduced system could be represented as an advection-diffusion-reaction system and the unknown dynamic can be learned by PARCv2.

For this case study, we trained PARCv2 using 100 simulation instances taken from DNS using the in-house code SCMITAR 3D \citep{Rai2015,Rai2017b,Rai2018}. Each simulation instance contains $15$ snapshots of state variable fields and velocity fields. The time interval between snapshots is kept constant at $\delta t = 0.17$ $(ns) $. All simulations are taken with the applied shock of $P_s = 9.5 (GPa)$. We also set aside 34 different simulation cases for the testing purpose. All the fields have the spatial dimension of $1.5 \times 2.25$ $(\mu m)$, tessellated using a uniform grid having the size of $128\times192$ \textit{(pixels)}. 

\subsection{Measuring Prediction Accuracy and Solution Quality}
\label{sec:metrics}
\paragraph{Burgers' Equation}
For the Burgers' equation problem, we measure the prediction accuracy by computing the average of the pixel-wise residual between prediction and ground truth velocity as: 
\begin{equation}
\label{eqn:residual_prediction_burgers}    
RMSE_u = \sqrt{\frac{1}{HW}\sum_{i=1}^{H} \sum_{j=1}^{W} |u_{ij} - \hat{u}_{ij}|^2 }
\end{equation}
Here $u_{ij}$ and $\hat{u}_{ij}$ are the velocity predicted by machine learning and the corresponding ground truth at the grid point $(i,j)$ on a $H \times W$ grid. The velocity value at each $(i,j)$ grid point is computed from the directional velocity as: 

\begin{equation}
\label{eqn:velocity_burgers}    
u_{ij} = \sqrt{u_x^2 + u_y^2},
\end{equation}
where $u_x$ and $u_y$ are the velocity values in horizontal and vertical directions, respectively. Meanwhile, the solution quality is measured by the average PDE residual of the governing PDE:
\begin{equation}
\label{eqn:solution_quality_burgers}    
|| \mathbf{f_u} || = \frac{1}{HW}\sum_{i=1}^{H} \sum_{j=1}^{W} \sqrt{(f_x)_{ij}^2 + (f_y)_{ij}^2},
\end{equation}

where $(f_x)_{ij}$ and $(f_x)_{ij}$ the PDE residual in horizontal and vertical directions at the grid $(i,j)$ and are computed as:
\begin{equation}
f_x =  \frac{\partial {u}_x}{\partial t} + u_x \frac{\partial u_x}{\partial x} + u_y \frac{\partial u_x}{\partial y} - \frac{1}{R}\left( \frac{\partial^2 u_x}{\partial x^2} + \frac{\partial^2 u_x}{\partial y^2}\right) \\
\end{equation}
\begin{equation}
f_y =  \frac{\partial {u}_y}{\partial t} + u_x \frac{\partial u_y}{\partial x} + u_y \frac{\partial u_y}{\partial y} - \frac{1}{R}\left( \frac{\partial^2 u_y}{\partial x^2} + \frac{\partial^2 u_y}{\partial y^2}\right)
\end{equation}

\paragraph{Navier-Stokes' Equation} In the case of Navier-Stokes equation, $RMSE_u$ is computed similarly with Burgers' equation using Eq. \ref{eqn:residual_prediction_burgers} and Eq. \ref{eqn:velocity_burgers}.

For measuring the solution quality, two physics-informed metrics were used: (1) the residual of the momentum PDE, and (2) the divergent-free condition of incompressible flow. The momentum residual is computed as:

\begin{equation}
\label{eqn:solution_quality_ns}    
|| \mathbf{f_u} ||= \frac{1}{HW}\sum_{i=1}^{H} \sum_{j=1}^{W} \sqrt{(f_x)_{ij}^2 + (f_y)_{ij}^2},
\end{equation}

where $(f_x)_{ij}^2$ and $(f_y)_{ij}^2$ are computed pixel-wise as:
\begin{equation}
f_x =  \frac{\partial {u}_x}{\partial t} + u_x \frac{\partial u_x}{\partial x} + u_y \frac{\partial u_x}{\partial y} - \frac{1}{\rho} \frac{\partial p}{\partial x} -  \frac{1}{Re} \left( \frac{\partial^2 u_x}{\partial x^2} + \frac{\partial^2 u_x}{\partial y^2}\right) \\
\end{equation}
\begin{equation}
f_y =  \frac{\partial {u}_y}{\partial t} + u_x \frac{\partial u_y}{\partial x} + u_y \frac{\partial u_y}{\partial y} - \frac{1}{\rho} \frac{\partial p}{\partial y} -  \frac{1}{Re} \left( \frac{\partial^2 u_y}{\partial x^2} + \frac{\partial^2 u_y}{\partial y^2}\right)
\end{equation}

The divergent-free condition error is computed as:
\begin{equation}
\varepsilon_{div} =  \frac{1}{HW}\sum_{i=1}^{H} \sum_{j=1}^{W} \left(\frac{\partial ({u}_x)_{ij}}{\partial x} + \frac{\partial ({u}_y)_{ij}}{\partial y}\right)\\
\end{equation}

\paragraph{Energy Localization in Energetic Materials}
We measure the prediction accuracy of PARCv2 and other models for the energetic materials problem by computing the RMSE of two state variables of interest, namely temperature and pressure. In a domain tessellated into $H\times W$ grid, the RMSE of the predicted temperature and pressure are computed as:

\begin{equation}
\label{eqn:acc_em_temp}    
RMSE_T = \sqrt{\frac{1}{HW}\sum_{i=1}^{H} \sum_{j=1}^{W} |T_{ij} - \hat{T}_{ij}|^2 }
\end{equation}

\begin{equation}
\label{eqn:acc_em_press}    
RMSE_P = \sqrt{\frac{1}{HW}\sum_{i=1}^{H} \sum_{j=1}^{W} |P_{ij} - \hat{P}_{ij}|^2 }
\end{equation}

Here $T_{ij}$ and $P_{ij}$ are the predicted temperature and pressure at the $(i,j)$ grid location, while the quantities with the hat ($\hat{}$) notation are ground truth derived from direct numerical simulation.

To assess the solution quality of machine learning prediction, we used the hotspot metrics introduced by \citet{Nguyen_single_void,Seshadri} for assessing the solution quality of predictions made by PARCv2 and other models. Particularly, we computed the RMSE of hotspot metrics, such that:

\begin{equation}
\label{eqn:epsilon_ths}    
\varepsilon_{T^{hs}} = \sqrt{\frac{1}{HW}\sum_{i=1}^{H} \sum_{j=1}^{W} |T^{hs} - \hat{T^{hs}}|^2 }
\end{equation}

\begin{equation}
\label{eqn:epsilon_ahs}    
\varepsilon_{A^{hs}} = \sqrt{\frac{1}{HW}\sum_{i=1}^{H} \sum_{j=1}^{W} |A^{hs} - \hat{A^{hs}}|^2 }
\end{equation}

\begin{equation}
\label{eqn:epsilon_tdot}    
\varepsilon_{\dot{T}^{hs}} = \sqrt{\frac{1}{HW}\sum_{i=1}^{H} \sum_{j=1}^{W} |\dot{T}^{hs} - \hat{\dot{T}}^{hs}|^2 }
\end{equation}

\begin{equation}
\label{eqn:epsilon_adot}    
\varepsilon_{\dot{A}^{hs}} = \sqrt{\frac{1}{HW}\sum_{i=1}^{H} \sum_{j=1}^{W} |\dot{A}^{hs} - \hat{\dot{A}}^{hs}|^2 }
\end{equation}

Here, the ${T}^{hs}$ is the \textit{average hotspot temperature}, ${A^{hs}}$ is the \textit{total hotspot area}, and $\dot{{T}}^{hs}$ and $\dot{A}^{hs}$ are their respective rates of change over time. These four QoIs are computed as:

\begin{equation}
\label{eqn:T_hs_bar}
    {T}^{hs}\left(t_k\right)=\frac{\sum_{i=1}^{H}\sum_{j=1}^{W}\left(T_{ij}^{hs}\left(t_k\right)A_{ij}^{hs}\left(t_k\right)\right)}{A^{hs}\left(t_k\right)},
\end{equation}

\begin{equation}
\label{eqn:A_hs_bar}
    {A^{hs}}(t_k)=\sum_{i=1}^{H}\sum_{j=1}^{W}A_{ij}^{hs}(t_k),
\end{equation}

\begin{equation}
\label{eqn:T_hs_dot}
  \dot{{T}}^{hs}(t_k)=\frac{{T}^{hs}\left(t_k\right)-{T}^{hs}\left(t_{k-1}\ \right)}{t_k-t_{k-1}},
\end{equation}

\begin{equation}
\label{eqn:A_hs_dot}
    \dot{A^{hs}}(t_k)=\frac{{A^{hs}}\left(t_j\right)-{A^{hs}}\left(t_{j-1}\right)}{t_k-t_{k-1}},
\end{equation}

\noindent where: 

\begin{equation}
\label{eqn:T_hs_compute}
T^{hs}_{ij}(t_k) = 
  \left\{
    \begin{array}{cl}
        T_{ij}(t_k) &\text{if}\quad T_{ij}(t_k)\geq875\ K,    \\
        0 & \text{otherwise,}   \\
     
    \end{array}
  \right.
\end{equation}

\begin{equation}
\label{eqn:A_hs_compute}
A^{hs}_{ij}(t_k)=
    \left\{
    \begin{array}{cl}
     A_{ij}(t_k) & \text{if}\quad T_{ij}(t_k)\geq875\ K,    \\
     0 & \text{otherwise.}   \\
    \end{array}
    \right.
\end{equation}

\noindent In Eqs. \cref{eqn:T_hs_bar,eqn:A_hs_bar,eqn:T_hs_dot,eqn:A_hs_dot,eqn:T_hs_compute,eqn:A_hs_compute}, the subscript $ij$ indicates quantities specific to the grid location $(i,j)$. The superscript $hs$ indicates quantities specific to hotspots. $A_{ij}$ is the area of the grid cell occupied in the hotspot region, which is constant since a uniform grid was employed. 

\subsection{PhyCRNet}
During our experiment, we kept the architecture of PhyCRNet intact with the one presented in the paper \citet{ren2022phycrnet}. Five convolution encoder blocks with filter sizes [4, 3, 4, 3, 4] and strides [2, 1, 2, 1, 2], one Conv-LSTM block with filter size 3 and stride 1, and one pixel shuffle layer with an upscaling factor of 8 as three 2x downsampling were performed during the encoding process. To eliminate the effect of number of parameters on prediction accuracy, we picked the number of channels to be [32, 32, 64, 64, 128, 128] for 2D Burger's equations and [128, 128, 256, 256, 512, 512] for Navier-Stokes equations to maintain parameter parity with PARCv2, yielding $\sim$1 million parameters for 2D Burgers and $\sim$20 million parameters for N-S equations. 


For PhyCRNet, we tested the performance of the model with only MSE data loss, with only PDE loss, and with a weighted combination of MSE and PDE loss on the 2D Burger's Equation problem, and we only used a weighted combination of MSE and PDE loss on the N-S equation experiment. We used Adam optimizer with an initial learning rate of $10^{-4}$ and for every 100 epochs we halved the learning rate with a minimal learning rate of $10^{-6}$. We stopped training when the test loss stopped improving for 50 epochs.

\subsection{FNO/PIFNO}
In our experiments, we maintained the same architecture of the Fourier Neural Operator (FNO) as the FNO-2d presented in the \citet{li2020fourier}: 2d FNO and an RNN structure in time. The 2d FNO first lifts the input to 32 channels for 2D Burgers (96 for N-S equations, 72 for EM) and then goes through 5 Fourier layers with 32 hidden channels (96 for N-S equations, 72 for EM), and lastly projects down to the number of channels in the output. The number of modes included in each Fourier layer is 15 for 2D Burgers and 24 for both N-S equations and EM, giving us $\sim$1 million parameters for 2D Burgers, $\sim$20 million parameters for N-S equations and $\sim$15 million parameters for EM, maintaining parameter parity with all other models presented in this work.

For FNO, we trained and tested the performance of the model with only MSE data loss in all three experiments. We used the Adam optimizer with an initial learning rate of $10^{-4}$, and for every 60 epochs, we halved the learning rate with a minimal learning rate of $10^{-6}$. We stopped training when the test loss stopped improving for 50 epochs. A weight decay of $10^{-5}$ was applied as regularization on both the 2D Burgers and EM experiments and no weight decay on N-S equations.

For PIFNO, we trained and tested the performance of the model with a weighted combination of MSE data loss and PDE solution quality in both 2D Bugers and N-S equations. We used the Adam optimizer with an initial learning rate of $10^{-4}$, and for every 60 epochs, we halved the learning rate with a minimal learning rate of $10^{-6}$. We stopped training when the test loss stopped improving for 50 epochs. No weight decay or any regularization was applied for PIFNO.

\subsection{Detailed Implementation of PARCv2}
The detailed implementation of PARCv2 and necessary data for reproducing the results can be found in \url{https://github.com/hphong1990/PARCv2}.

For PARCv2, we used Adam optimizer with a learning rate of $10^{-5}$. As mentioned before, we trained PARCv2 using a two-step training. The differentiator was trained first for 500 epochs. Consequently, the differentiator is frozen and the integrator was trained for another 500 epochs. For the PARC setting without the data-driven integrator, only the differentiator was trained. No weight decay or any extra regularization was applied.

\subsection{Comparison of Computational Cost}
Table  \ref{tab:runtime} reports the computation cost of PARCv2 and other baselines compared to DNS. All baselines were tested on an NVIDIA A100 Graphic Computing Unit (GPU). DNS was conducted on a high-performance computing (HPC) system with Intel® Xeon E5-2699v4 CPU @ 2.80 GHz (1,320 processors) and NVIDIA P100 GPUs accelerator.

\begin{table}[tbh!]
\caption{Computation cost comparison}
\label{tab:runtime}
\vskip 0.15in
\begin{center}
\begin{small}
\begin{sc}
\begin{tabular}{lccc}
\toprule
                & Burgers & Naiver-Stokes & EM\\
\midrule
PhyCRNet   & 0.264 (s)& 0.713 (s)& $--$\\
FNO	&0.433 (s)	&0.939 (s)	&0.153 (s)\\
PIFNO	&0.433 (s)	&0.939 (s)	&-- \\
PARCv2	&0.445 (s)	&0.278 (s)	&0.820(s) \\
PARC (num. int.)	&0.122 (s)	&0.127 (s)	&0.089 (s)\\
PARC (NN. int.)	&0.158 (s)	&0.090 (s)	&0.149 (s)\\
DNS	&493.2 (s)	&$~$ 1-2 (hours)	&$~$2-3 (hours)\\

\bottomrule
\end{tabular}
\end{sc}
\end{small}
\end{center}
\vskip -0.1in
\end{table}


\end{document}